\documentclass[letterpaper, 10 pt, journal, twoside]{ieeetran}
\usepackage{graphicx}
\usepackage{mathtools, nccmath}
\usepackage{multirow}
\usepackage{algorithm,algpseudocode}
\usepackage{booktabs}
\usepackage{fancyhdr}   
\usepackage{gensymb}
\usepackage{comment}
\usepackage{notoccite}
\usepackage{amsfonts}
\usepackage{bm}

\newcommand{\uvec}[1]{\boldsymbol{\hat{\textbf{#1}}}}
\newcommand{\cmmnt}[1]{}

\usepackage[font=footnotesize]{caption}
\usepackage{cite}
\usepackage{xcolor}

\usepackage[letterpaper, top=19.2mm, bottom=19.2mm, left=19.2mm, right=19.2mm]{geometry}

\usepackage{titlesec}
\titlespacing{\section}{0pt}{*0.2}{*0.2}
\titlespacing{\subsection}{5pt}{*0.15}{*0.15}
\IEEEoverridecommandlockouts                   

\author{Rhys Newbury$^{1*}$, Kerry He$^{1*}$, Akansel Cosgun$^{1}$ and Tom Drummond$^{1}$%

\thanks{*Equal contribution}
\thanks{$^{1}$Monash University, Australia
        {Email: \tt\footnotesize rhys.newbury@monash.edu}}%
\thanks{Digital Object Identifier (DOI): 10.1109/LRA.2021.3068122}
\thanks{\textcopyright2021 IEEE. Personal use of this material is permitted.  Permission from IEEE must be obtained for all other uses, in any current or future media, including reprinting/republishing this material for advertising or promotional purposes, creating new collective works, for resale or redistribution to servers or lists, or reuse of any copyrighted component of this work in other works.}
}

\title{\vspace*{6mm}Learning to Place Objects onto Flat Surfaces in Upright Orientations}

\begin{document}


\maketitle
\thispagestyle{empty}

\begin{abstract}
We study the problem of placing a grasped object on an empty flat surface in an upright orientation, such as placing a cup on its bottom rather than on its side. We aim to find the required object rotation such that when the gripper is opened after the object makes contact with the surface, the object would be stably placed in the upright orientation. We iteratively use two neural networks. At every iteration, we use a convolutional neural network to estimate the required object rotation, which is executed by the robot, and then a separate convolutional neural network to estimate the quality of a placement in its current orientation. Our approach places previously unseen objects in upright orientations with a success rate of 98.1\% in free space and 90.3\% with a simulated robotic arm, using a dataset of 50 everyday objects in simulation experiments. Real-world experiments were performed, which achieved an 88.0\% success rate, which serves as a proof-of-concept for direct sim-to-real transfer.
\end{abstract}

\section{Introduction}
\label{sec:introduction}

Everyday objects are usually placed in specific orientations that are convenient to humans. For example, a cup is designed to be placed on its bottom rather than on its side. Placing objects down correctly in semantically preferable orientations is a fundamental skill for service robots. For example, a robot unpacking the dishwasher should place plates, glasses, and bowls on shelves in certain orientations. Research in robotic manipulation over the past decades has mostly focused on how to pick up objects~\cite{DataDrivenGrasp}, with recent works utilizing the advances in deep learning~\cite{lenz2013deep,mahler2017dexnet,pinto2015supersizing,morrison2018closing}. However, after it has been grasped what to do with the object has largely been overlooked in the field. A common practice in pick-and-place robotic manipulation scenarios is to drop the object at a height without any consideration to its resulting pose~\cite{mahler2017dexnet, morrison2018closing, 2018Garrett}. Only a handful of researchers have studied how to place the grasped object down~\cite{Jiang2011, harada2012, haustein2019object,Fu2008}. Furthermore, no work to our knowledge has leveraged deep learning for the object placement problem.

\begin{figure}[ht!]%
    \centering
    \includegraphics[width=0.99\linewidth]{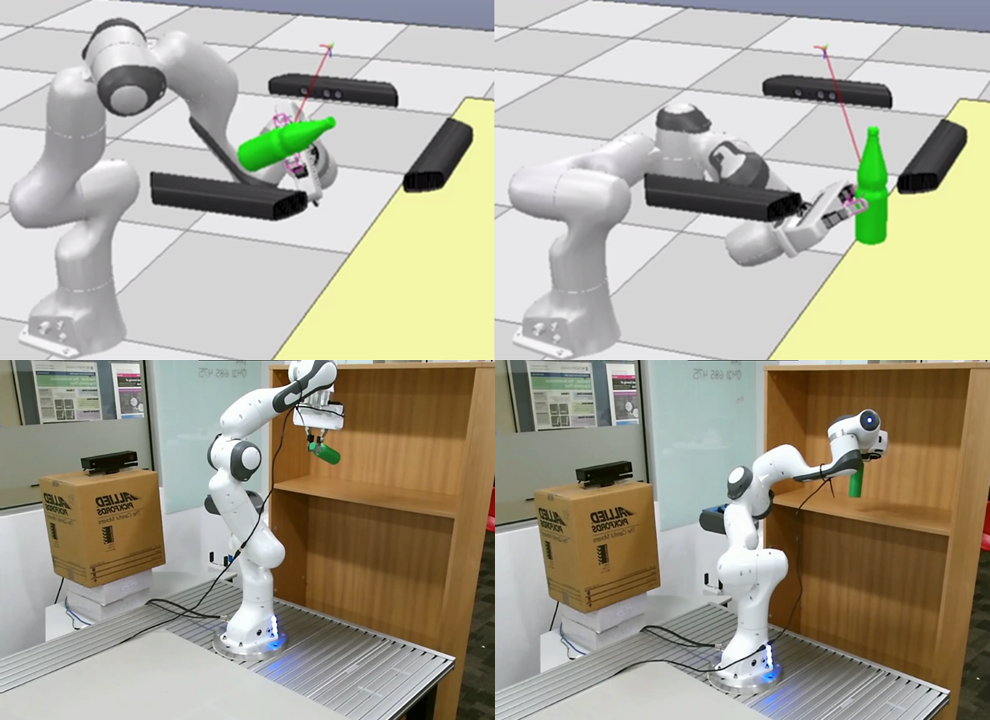}
    \caption{Object placement of a coke bottle upright on a flat surface in simulation (top) and of a 3D-printed model using a real robot (bottom). The robot starts with a random grasp pose (left). Our approach computes a rotation required to bring the object to an upright orientation (right).}
    \label{fig:intro}
\end{figure}

Humans usually associate an upright orientation with objects, placing them in a way that they are most commonly seen in our surroundings~\cite{Fu2008}. In this paper, we study the problem of placing a grasped object down on an empty flat surface in the upright orientation. We consider the solution to the problem as finding the required object rotation such that when the gripper is opened, the object comes to rest in the upright orientation under the influence of gravitational forces. We propose two convolutional neural networks, Placement Rotation Convolutional Neural Network (PR-CNN) and Placement Quality Convolutional Neural Network (PQ-CNN), used in an iterative algorithm until a satisfactory solution is found. We train both networks in simulation using 45 randomly selected object models of typical household objects. We report the experimental results on the remaining five previously unseen objects starting from random object orientations in the robotic gripper.

The contributions of this paper is twofold:
\begin{itemize}
\setlength{\itemsep}{0pt}
\setlength{\parskip}{0pt}
\setlength{\parsep}{0pt}
    \item An iterative, learning-based approach to placing previously unseen objects from input depth images, without access to object class or models.
    \item A proof-of-concept implementation on a robotic system, demonstrating the feasibility of direct sim-to-real transfer.
\end{itemize}

The organization of the paper is as follows.  We review the relevant literature in Section~\ref{sec:related_work} and describe the problem in Section~\ref{sec:problem_description}. We present our iterative placement method in Section~\ref{sec:approach}. We investigate placement in free space in Section~\ref{sec:experimental_setup}, with a simulated robot in Section~\ref{sec:simulated_robot} and a real robot in Section~\ref{sec:real_robot}, before concluding in Section \ref{sec:conclusion}.

\section{Related Work}
\label{sec:related_work}


In one of the earliest implementations of robotic object placement, Edsinger~\cite{EdSigner} use a compliant robotic arm to place an object onto a shelf by moving the arm to a fixed configuration and then lowering the end-effector using force control, hence utilizing contact with the environment. However, this approach assumes that the pose of the object in the gripper is known, which is unavailable for unknown objects. Since then, many researchers have approached the placement problem analytically, attempting to find flat features on the object and the surface on which the object can be placed~\cite{baumgartl2013, baumgartl2014c,harada2012, HARADA20141463,haustein2019object}. Baumgart~\cite{baumgartl2013,baumgartl2014c} find stable poses of the object analytically by finding a point of first object contact, followed by rotating the object such that additional contact points are found. Their approach runs in real time, but does not take into account object semantics for determining placement orientations. Harada~\cite{harada2012,HARADA20141463} matches planar surface patches on the object with planar surface patches in the environment, which allows finding placements on large, flat surfaces, but also less obvious placements such as a mug hanging on a flat bar. Their approach, however, requires the 3D model of the object and its pose. Haustein~\cite{haustein2019object} presents a similar approach and uses Monte Carlo Tree Search to optimize motion planning to reach the stable pose.

The majority of recent approaches in robotic manipulation are driven by machine learning approaches, and object placement is no exception. Fu~\cite{Fu2008} use hand-chosen features to find the upright orientation of man-made objects. The approach presented by Jiang~\cite{Jiang2011,Jiang2012} uses learning on hand-designed features and successfully places known objects in stably $98\%$ of the time and new objects $82\%$ of the time. Paolini~\cite{Paolini2014} estimates the probability of a successful placement, then attempts to solve for the most likely placement location, given a grasped object. Manuelli~\cite{manuelli2019kpam} develops a placement algorithm based on keypoint estimators, which applies geometric constraints on the keypoints to achieve category-level placement. This work is extended to use shape completion based on dense point clouds~\cite{gao2019kpamsc}. Mitash~\cite{mitash2020task} demonstrates successful placement of previously unknown objects into constrained spaces by using a possibilistic representation to model the unobserved parts of the object. They also demonstrate a transfer of the object between two robot manipulators, from a vacuum-based to a parallel jaw gripper.

While the majority of the surveyed papers make use of 3D object models, similar to \cite{Jiang2011,Jiang2012,mitash2020task}, we assume that we do not have access to the object model or class during run-time and must solve the placement problem utilizing only sensor readings from RGBD cameras. Our work is different from existing literature in two aspects: 1) The concept of upright orientations, which embodies human preferences, has been first explored by Fu~\cite{Fu2008} at a theoretical level given object models. Our work is novel in combining this theory with the robotic placement problem. 2) We propose a novel hardware-in-the-loop iterative approach, where the robot applies rotations to the object, and the new observations from the sensors are passed to the network until convergence to the upright object orientation is achieved.

\section{Problem Description}
\label{sec:problem_description}

\begin{figure}[t!]
\centering
\includegraphics[width=0.6\linewidth]{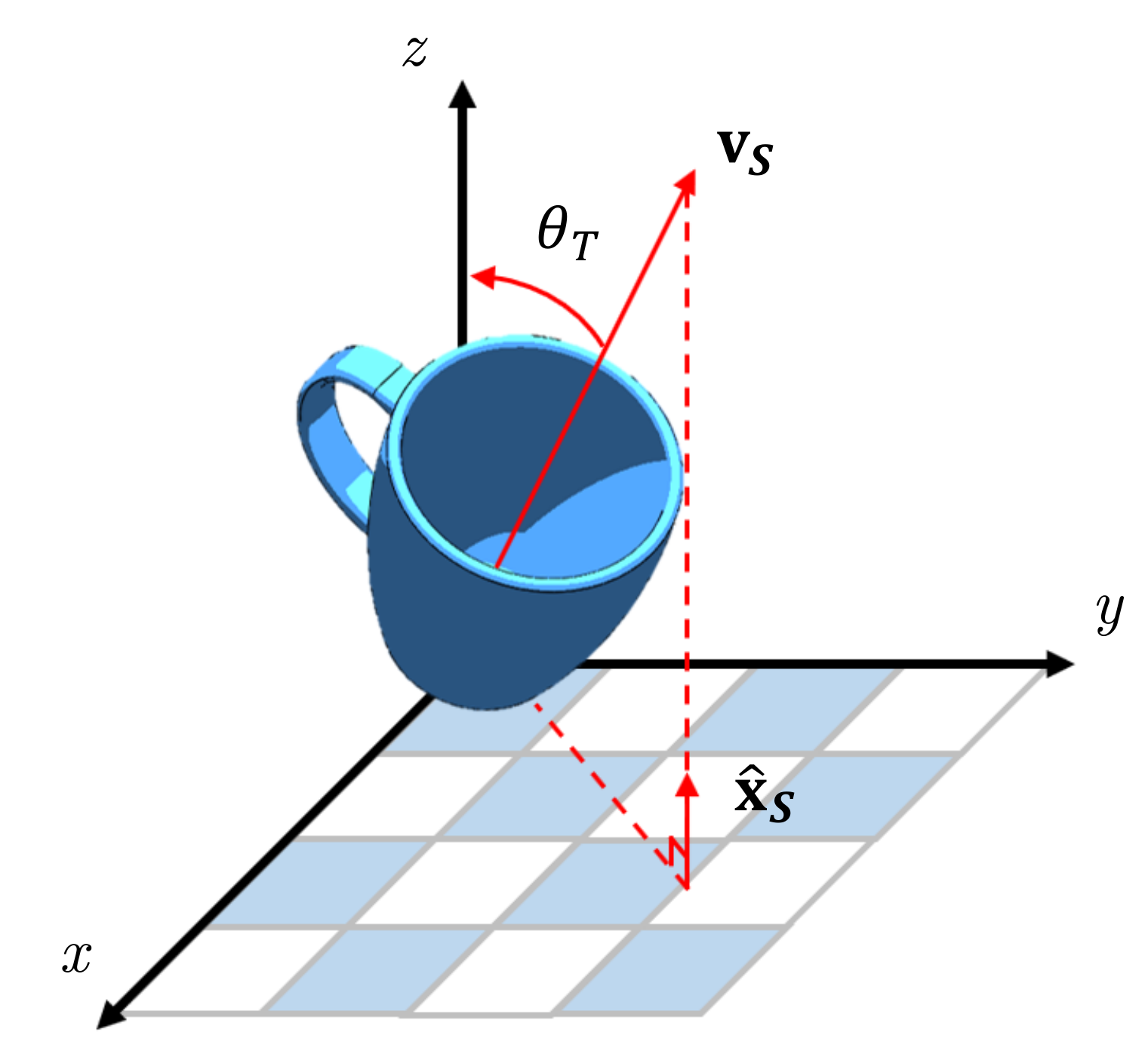}
\caption{We define the placement surface as the infinite $xy$-plane, stable axis to the placement plane ($\uvec{x}_{\bm{S}}$) and upright vector for each object ($\uvec{v}_{\bm{S}}$) which encodes possible upright orientations. $\theta_{T}$ is the shortest angle to rotate the object to an upright orientation.}
\label{fig:3d_diagram}
\end{figure}

Fig.~\ref{fig:3d_diagram} illustrates the axes and angles we define for an object. The coordinate frame defined by the axes ($x,y,z$) represents an arbitrary and fixed world frame, in which gravity is acting in the -$z$ direction. We define the stable axis $\uvec{x}_{\bm{S}}$ as the orthogonal unit vector to the surface on which objects are placed. The placement surface is set as the infinite and uncluttered $xy$-plane; hence we set $\uvec{x}_{\bm{S}}$ to be in the direction of the $z$-axis.  We assign an upright vector $\uvec{v}_{\bm{S}}$ attached to an object, such that $\uvec{v}_{\bm{S}}=\uvec{x}_{\bm{S}}$ when the object is in its upright orientation. An important property of object placement on an uncluttered infinite plane is that the placement is independent of any rotation about the stable axis $\uvec{x}_{\bm{S}}$. To illustrate this, any rotation of the mug shown in Fig.~\ref{fig:3d_diagram} about $\uvec{x}_{\bm{S}}$ can be reframed as a rotation of the global reference frame about the same axis. Because of this, there exists a continuous set of rotations $\textbf{R}\in SO(3)$ that can orient an object to an upright orientation. We uniquely define the ground truth rotation $R_T$ as the shortest possible rotation required to move the object from its current orientation to an upright orientation. The ground truth rotation $R_T$ can therefore be found as the transformation required to rotate the unit vector $\uvec{v}_S$ to $\uvec{x}_S$. This rotation is calculated most easily using axis-angle representation, which is defined by an angle $\theta$ rotated about an axis $\bm{n}$. The ground truth rotation angle $\theta_{T}\in[-\pi, \pi]$ is the angle between unit vectors $\uvec{v}_S$ and $\uvec{x}_S$ and can be found by:

\begin{equation}
\label{dot}
    \theta_{T}=\arccos{(\uvec{x}_S\cdot \uvec{v}_S)}
\end{equation}

The ground truth axis $\bm{n}_{\bm{T}}$ is defined as the axis perpendicular to both $\uvec{v}_S$ and $\uvec{x}_S$ and can be found by:

\begin{equation}
\label{cross}
    \bm{n_{T}}=\uvec{x}_S\times \uvec{v}_S
\end{equation}

We consider the robotic placement problem as follows. The robot starts with an object already in hand, and the task is to place the object down successfully. A solution to the problem is a proposed object rotation that would result in a particular object orientation (the upright orientation in this case). We assume that the robot has no a priori knowledge about the object class, 3D model, or the upright orientation, however, it has access to depth cameras and force sensing. Successful placement onto the tabletop requires that the object is stable and in the upright orientation under gravitational and contact forces after release.

\section{Iterative Placement with Quality}
\label{sec:approach}



Our approach draws inspiration from the Iterative Error Feedback idea proposed by Carreira~\cite{carreira2016human}, which progressively changes an initial solution by feeding back error predictions. We also draw inspiration from actor-critic networks, where one network outputs an action for the system to execute, and the other network determines the quality of the action~\cite{actorcritic}. We propose an iterative, learning-based approach to robotic object placement. We propose two neural networks which both take as input depth images from multiple viewpoints:

\begin{itemize}
    \item Placement Rotation Convolutional Neural Network (PR-CNN): Outputs the rotation that transforms the object to the upright orientation.
    \item Placement Quality Convolutional Neural Network (PQ-CNN): Estimates the confidence level that the object would be come to rest in its upright orientation if it is placed in its current orientation.
\end{itemize}

\begin{algorithm}[H]
\caption{Proposed placement algorithm}
\label{alg:proposed}
\begin{algorithmic}[1]
    \Function{RotateObject}{}
        \For{$i = 1 \to max\_restart$}    
            \For{$i = 1 \to max\_iter$}
                \State {$img \gets take\_depth\_image()$}
                \State {$rotation[i] \gets \textbf{PR-CNN}(img)$}
                \State {$rotate\_ee\_by(rotation[i])$}
                \State {$img \gets take\_depth\_image()$}
                \State {$orientation[i] \gets get\_ee\_rotation()$}
                \State {$quality[i] \gets \textbf{PQ-CNN}(img)$}
                
                \If{${1 - quality[i]} <\epsilon_1$}
                    \If{$|rotation[i]| < \epsilon_2$}
                        \State {\textbf{goto 18}}
                    \EndIf
                \EndIf
            \EndFor   
            \State {$rotate\_ee\_by(random\_rotation())$}
        \EndFor
        \State {$max\_index \gets arg\_max(quality)$}
        \State {$rotate\_ee\_to(orientation[max\_index])$}
    \EndFunction
\end{algorithmic}
\end{algorithm}


The pseudocode can be seen in Algorithm~\ref{alg:proposed}. At every iteration, the PR-CNN proposes an object rotation to achieve the upright orientation, which the robot executes. Then PQ-CNN estimates the quality of the proposed object orientation. The algorithm stops if PQ-CNN predicts that the orientation would be upright if placed in the current orientation, and PR-CNN's output is the rotational identity which suggests that the object orientation has converged. If this does not occur within a number of iterations $max\_iter$, the algorithm restarts from a new random object orientation. If the maximum number of restarts $max\_restart$ is reached without a satisfactory solution, the rotation that yields the maximum predicted quality is chosen among all iterations. The iterative approach helps in getting more observations from the object and correcting the errors of PR-CNN. Quality estimation provided by PQ-CNN is useful to identify when PR-CNN converges to an upside-down orientation, which is a common failure mode when objects lack noticeable features distinguishing the upright and upside-down orientations.



\subsection{Placement Rotation CNN (PR-CNN)}
\label{subsec:prcnn}

We aim to learn the required rotation applied to the object that would result in an upright orientation. The ground truth rotation $R_T$ is obtained analytically using the methodology outlined in Section~\ref{sec:problem_description}.





\begin{figure}[ht!]
\centering
\includegraphics[width=0.95\linewidth]{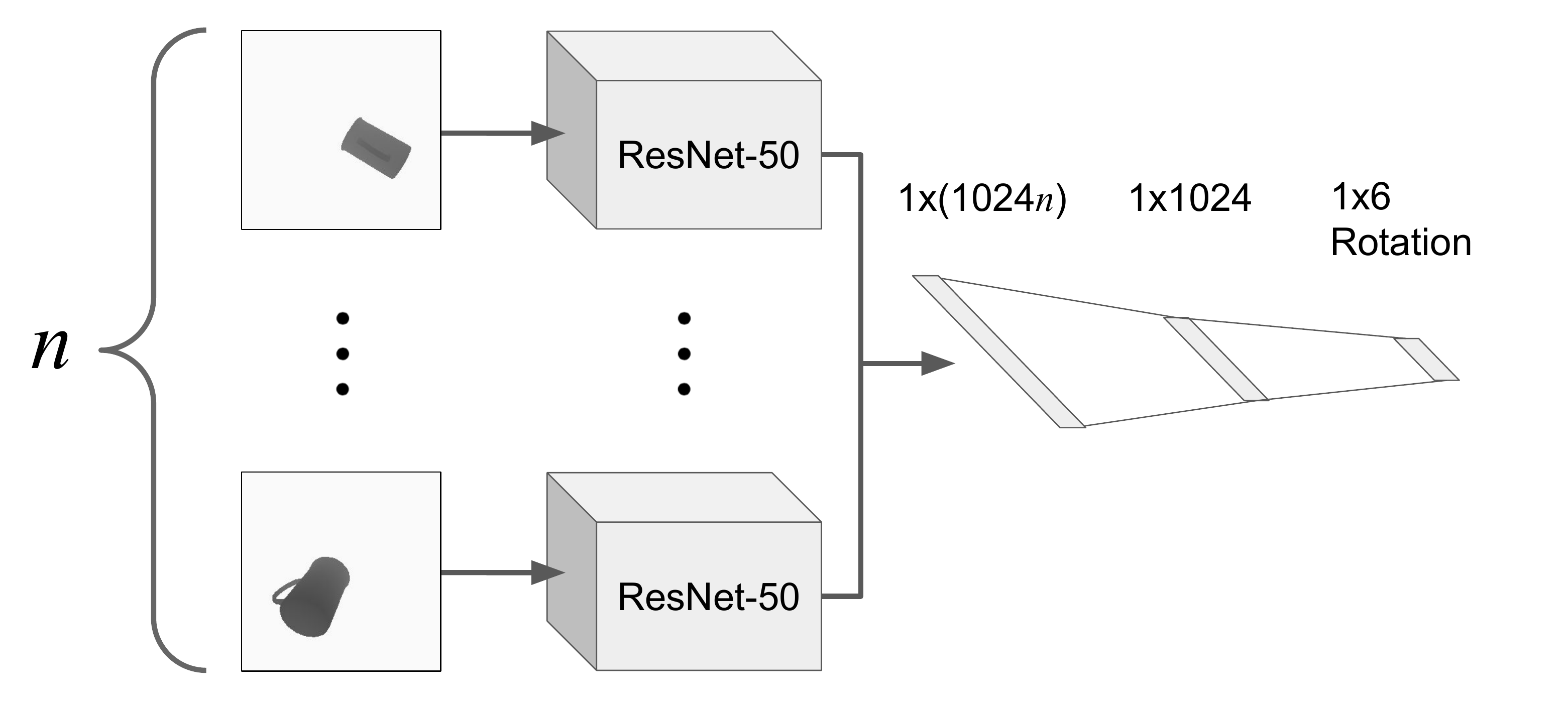}
\caption{We train a convolutional neural network, with shared weights between each of the $n$ depth images. These are then passed through linear layers to get to the desired number of outputs}
\label{fig:CNN}
\end{figure}

PR-CNN has an architecture based on ResNet-50~\cite{he2015deep} and is pre-trained on ImageNet~\cite{Deng09imagenet}. PR-CNN takes as input $n$ $64\times64$ depth images of the object, which has depth values normalized between $[-0.5, 0.5]$. The network is not specific to the number or pose of cameras and may vary between implementations discussed in this paper. We train a single CNN with shared weights, such that each depth image is passed independently to the network. The $n$ outputs of size $1024$ are then concatenated into a single output. This output is then used in two linear layers reducing the output size to $1024$ and $6$, respectively. This network structure (shown in Fig.\ref{fig:CNN}) draws inspiration from multi-viewpoint pose estimation~\cite{openai2019learning}. The six-dimensional output corresponds to the six-dimensional representation of $SO(3)$ rotations as proposed by Zhou~\cite{zhou2018continuity}. This represents the first two columns of a rotation matrix, which is then converted to a full rotation matrix using the Gram-Schmidt-like process described by Zhou~\cite{zhou2018continuity}. This representation of rotations is continuous, which can allow for smaller approximation errors~\cite{Chen2013TheCA}.

Furthermore, we use a similar loss function to Zhou~\cite{zhou2018continuity}, which is the geodesic distance between the output ($R_s$) and the ground truth ($R_T$):

\begin{equation}
\label{geodesic}
    \mathcal{L}_{geodesic} = \arccos{\left(\frac{tr(R_sR_T^{-1})-1}{2}\right)}
\end{equation}

where $tr(R)$ is the matrix trace operator. For a rotation matrix, this is defined as:

\begin{equation}
\label{trace}
    tr(R)=R_{00}+R_{11}+R_{22}
\end{equation}

\subsection{Placement Quality CNN (PQ-CNN)}

We learn the estimated quality of an object being placed in its current orientation, which is defined as the confidence that the object will come to rest in the upright orientation if released in the current orientation. The object quality that we are estimating is based on the assumption of placing on a flat surface. The ground truth binary label is obtained by Bullet Physics Simulation~\cite{coumans2010bullet}. The procedure to collect data is described in Section~\ref{sec:data_collection}. By learning the estimated quality of an object in its current pose, we can avoid cases where PR-CNN proposes incorrect rotations. However, this network can be used independently of PR-CNN to estimate the quality of placements using a traditional approach to find the upright pose.

Human preferred upright orientations of objects are often defined by the personal design and preference of humans, meaning an analytical approach may be difficult to generalize between different classes of objects. Hence, this motivates the use of a CNN, which can be trained using user-defined upright orientations.

PQ-CNN uses the same network structure as PR-CNN (Section~\ref{subsec:prcnn}). The weights are not shared between the ResNet-50 architectures as the networks are learning two different criteria, and we did not want to bias the networks to learn the same patterns. Modifications were made to the final layer to output a single number with sigmoid activation. We use a binary cross-entropy loss function for PQ-CNN.





\section{Experiments Without a Robot}
\label{sec:experimental_setup}

\subsection{Object Models}
\label{subsec:objmodel}

We use a dataset of $50$ everyday objects, each with 3D meshes of each object. These object models are only needed for data collection and are not used during the execution of our approach. The simulated renderings of all $50$ objects can be seen in Fig 4. We picked the objects such that each object had a well-defined, single upright vector. We avoided symmetrical objects such as cans and boxes due to the existence of multiple stable orientations, which are impossible for a depth camera to differentiate between, leading to an ambiguous problem. We also avoided objects without a ``natural base" such as spherical objects, spoons, or toothbrushes, as these objects were not designed to stand in an upright orientation~\cite{Fu2008}. We manually label each object with an upright orientation. We use a scene where three depth cameras were positioned orthogonally from each other, to the front, left, and right of the gripper, at a $25$cm radius around the point $p_c$ at which objects and the gripper interact (as seen in Fig.~\ref{fig:intro}).

\begin{figure}[ht!]
\centering
\includegraphics[trim=0.5cm 0.6cm 0.55cm 0.4cm, clip, width=1.0\linewidth]{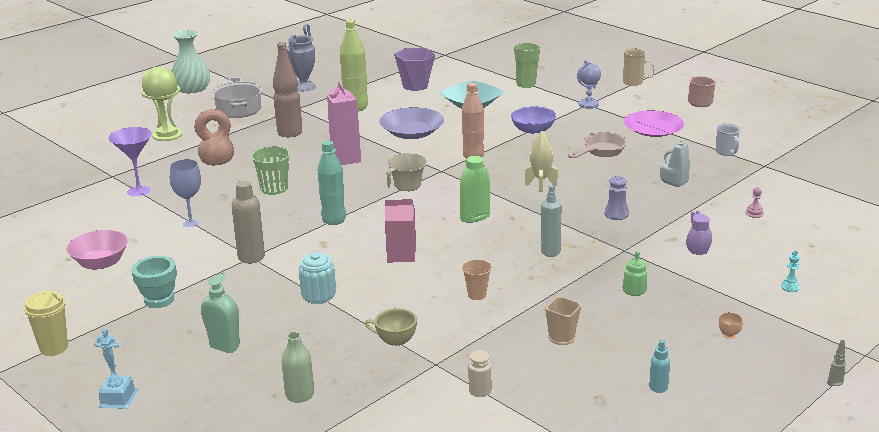}
\caption{The $50$ object models that were chosen to train and evaluate our models. Object models are depicted in their upright orientations which are annotated manually. Test sets were created by randomly selecting 5 objects as test objects to perform evaluation on, while the remaining 45 objects are used to train the networks.}
\label{fig:objects}
\end{figure}

\subsection{Data Collection}
\label{sec:data_collection}

We used the PyRep toolkit~\cite{pyrep} for the simulation environment. PR-CNN and PQ-CNN need separate but overlapping datasets. We collected two datasets, both with $100,000$ data points. To collect a data point, for PR-CNN, we randomly pick an object from the $50$ objects and randomly sample an orientation with a slight positional variation around $p_c$. We collect the three depth camera images and the ground truth rotation that would rotate the object to the upright orientation. We save the binary label for PQ-CNN, indicating whether the object would be in an upright placement or not when the object is dropped at the lowest possible height from the placement surface in the sampled orientation (obtained by physics simulation). As PQ-CNN aims to solve a binary classification task, we need a balanced dataset for both classes. As random orientations are more likely to unsuccessfully come to rest in the upright orientation, we sample additional data points with slight variations to the upright orientation. Furthermore, we use an equal amount of successful and unsuccessful data points for training PQ-CNN.

\subsection{Methods}

\begin{itemize}
    \item\textbf{Baseline:} We combine point clouds from $6$ orthogonal rotations (analogous to 6 sides of a cube) and estimate each point's normal vector. The object is then placed such that the average normal vector of the largest flat plane is perpendicular to the table surface. Open3D~\cite{Zhou2018} is used for point cloud operations.
    
    \item\textbf{Single pass (SP):} A single pass of PR-CNN is used to determine the object rotation.
    
    \item\textbf{Iterative (ITR):} PR-CNN is run iteratively until the identity rotation is achieved or a maximum number iterations (15 in this case) has been reached.
    
    \item\textbf{Iterative with Quality (ITR-Q):} Our full approach combining PR-CNN and PQ-CNN, as detailed in Section~\ref{sec:approach}.
    
\end{itemize}

\subsection{Metrics}

\begin{itemize}
    \item \textbf{Success Rate:} The percentage of placements where the steady-state object orientation is within $\pm15\degree$ of the desired ground truth orientation.
    \item \textbf{Stability Rate:} The percentage of placements where the object stays stationary for a minute and the final object orientation is within $\pm15\degree$ of the initial placement orientation. This metric was adopted from~\cite{Jiang2012}.
    \item \textbf{Angular Error:} The average angle difference between the upright vector  $\uvec{v}_S'$ and the stable axis $\uvec{x}_S$.
\end{itemize}





\subsection{Experimental Procedure}
PR-CNN was trained for $150$ epochs, and the epoch with the highest \textbf{ITR} success rate was selected to present results. For PR-CNN, we did not use a validation set, however, we validated the model on the test objects by using physics to simulate $250$ placements from random initial orientations of randomly selected objects every $25$ epochs. PQ-CNN was trained for $15$ epochs, and the highest validation accuracy was used during the evaluation of \textbf{ITR-Q}.


\subsection{Design Parameters}
\label{sec:design-parameteres}

To select the design parameters used to evaluate the full \textbf{ITR-Q} approach, we performed additional studies on PR-CNN to investigate the effect of pre-training, shared weights, output angle representation, and number of cameras. Unless otherwise specified, we use pre-training, shared weights, 6D angular representation, and 3 orthogonal cameras for training the PR-CNN. To attribute any performance differences to PR-CNN, success rates are obtained using only \textbf{ITR} approaches. For these experiments, we trained and evaluated the networks on all 50 object models. The experiments were completed using a NVIDIA GeForce GTX 1080. The average network inference time was $11.89$ms for PR-CNN (three depth image inputs).

\textbf{Network Architecture:} We investigate the effect of pre-training and shared weights on the performance of the network. Each network uses ResNet-50\cite{he2015deep} as the backbone. We consider two alternatives, firstly, training a singular network with three input channels, where each channel represents a depth image from a different viewpoint. Secondly, an architecture using shared weights as described in Section~\ref{subsec:prcnn}. The results are shown in Table~\ref{table:architecture}. The effect of pre-training is particularly noticeable from the results, which show an increase in success rate from $54.8\%$ to $98.4\%$ for \textbf{ITR}. Despite being pre-trained on an unrelated dataset (ImageNet\cite{Deng09imagenet}) consisting of 3 channel RGB images, the network trains faster and converges to a lower loss. We also observe a performance increase of $8.8$ percentage points in using the architecture with shared weights using \textbf{ITR}. For the rest of this paper, we use pre-training along with the architecture with shared weights.

\begin{table}[t!]
\vspace{0.25cm}
\centering
\begin{tabular}{@{}lc@{}}
\toprule
 & \textbf{ITR (Success Rate \%)} \\ \midrule
\multicolumn{1}{l|}{\textbf{ResNet-50 SW}} & 54.8 \\
\multicolumn{1}{l|}{\textbf{ResNet-50 PT}} & 89.6 \\
\multicolumn{1}{l|}{\textbf{ResNet-50 PT SW}} & \textbf{98.4} \\ \bottomrule
\end{tabular}
\caption{Comparison of network architectures using pre-training (PT) and shared weights (SW). Pre-training yields an improvement in success rate. The architecture using three single-channel input CNN's in parallel with shared weights is shown to outperform the alternative architecture using a single three-channel input CNN.}
\label{table:architecture}
\end{table}

\textbf{Number of Cameras:} We analyze the effect of using different quantities of cameras on the placement performance. All cameras are orthogonal to each other as described in Section~\ref{subsec:objmodel}, with additional cameras being added to the left, front, right, and back of the object in the order of increasing cameras. The results are shown in Table~\ref{table:cameras}. There is a noticeable performance increase when going from one to two cameras. However, the benefit of additional cameras plateaus at three cameras. Although three cameras were used in this paper, these results suggest that our approach can be flexibly transitioned to use only one or two cameras without a major drop in performance. This allows our approach to be more applicable to the real world, where access to three cameras may not be possible or practical. Our real-world robotic experiments are conducted with a single camera.

\begin{table}[h!]
\centering
\begin{tabular}{@{}lc@{}}
\toprule
 & \multicolumn{1}{l}{\textbf{ITR (Success rate \%)}} \\ \midrule
\multicolumn{1}{l|}{\textbf{1 Camera}} & 85.2 \\
\multicolumn{1}{l|}{\textbf{2 Cameras}} & 96.8 \\
\multicolumn{1}{l|}{\textbf{3 Cameras}} & \textbf{98.4} \\
\multicolumn{1}{l|}{\textbf{4 Cameras}} & 96.4 \\ \bottomrule
\end{tabular}
\caption{Comparison of different quantities of cameras used as inputs into PR-CNN. The setup using three cameras performed the best out of all \textbf{ITR} experiments.}
\label{table:cameras}
\end{table}

\textbf{Angular Representation:} We compare the effect of using different angular representations as the output of PR-CNN. For all angular representations, we use the same geodesic loss function shown in Eq.~\ref{geodesic}. The results are shown in Table~\ref{table:angle}. The Euler angle representation performed worst among all representations, likely due to discontinuity issues. The 6D angular representation~\cite{zhou2018continuity} slightly outperformed the quaternion representation by $2.0$ percentage points. While both the quaternion and 6D representations are viable options for the network, we opt to use the 6D representations for the rest of this paper.

\begin{table}[h!]
\centering
\begin{tabular}{@{}lc@{}}
\toprule
 & \textbf{ITR (Success rate \%)} \\ \midrule
\multicolumn{1}{l|}{\textbf{Euler}} & 32.4 \\
\multicolumn{1}{l|}{\textbf{Quaternion}} & 96.4 \\
\multicolumn{1}{l|}{\textbf{6D~\cite{zhou2018continuity}}} & \textbf{98.4} \\ \bottomrule
\end{tabular}
\caption{Comparison of different angular representations on network performance. The Euler representation performed the worst, while the 6D representation yielded the highest \textbf{ITR} success rate.}
\label{table:angle}
\end{table}

\subsection{Performance on Unseen Objects}
\label{sec:unseen}
To evaluate the generalizability of the networks, five test sets were created by randomly selecting $5$ non-overlapping objects out of the $50$. We trained PQ-CNN and PR-CNN on the remaining $45$ objects. The aggregate results on all the test sets is shown in Table~\ref{table:mainresults}.

\begin{table}[h!]
\centering
\begin{tabular*}{\columnwidth}{@{\extracolsep{\stretch{1}}}*{5}{c}@{}}
\toprule
 & \textbf{\begin{tabular}[c]{@{}c@{}}Success\\ Rate (\%)\end{tabular}} & \textbf{\begin{tabular}[c]{@{}c@{}}Stability Rate\\ (\%)\end{tabular}} & \textbf{\begin{tabular}[c]{@{}c@{}}Angular\\ Error ($\degree$)\end{tabular}} \\ \midrule
\multicolumn{1}{l|}{\textbf{Baseline}} & 54.0 $\pm$ 10.0 & 96.7 $\pm$ 4.2 & 47.7 $\pm$ 11.5 \\
\multicolumn{1}{l|}{\textbf{SP}} & 84.4 $\pm$ 8.9 & 89.9 $\pm$ 5.8 & 22.8 $\pm$ 13.1 \\
\multicolumn{1}{l|}{\textbf{ITR}} & 96.1 $\pm$ 4.5 & 98.3 $\pm$ 1.6 & 8.0 $\pm$ 7.1 \\
\multicolumn{1}{l|}{\textbf{ITR-Q}} & \textbf{98.1} $\pm$ 1.9 & \textbf{99.3} $\pm$ 1.1 & \textbf{5.2} $\pm$ 2.5 \\ \bottomrule
\end{tabular*}
\caption{Performance of different placement methods averaged over five different sets of 5 previously unseen objects randomly chosen from the set of 50 objects. \textbf{SP}, \textbf{ITR} and \textbf{ITR-Q} were trained on the remaining 45 known objects.  Our full approach \textbf{ITR-Q} outperforms all other  methods  in  all  metrics.  Standard  deviations  are  between  the  metrics obtained from the five test sets.}
\label{table:mainresults}
\end{table}

\textbf{SP}, \textbf{ITR} and \textbf{ITR-Q} all outperformed the baseline. Although the baseline had a high stability rate of $96.7\%$, it only places objects in upright orientations only $54.0\%$ of the time. This suggests that the largest flat surface of an object often corresponds to a stable pose but does not always correspond to a semantically preferred orientation.

The proposed \textbf{ITR-Q} algorithm performed the best in all metrics, achieving a success rate of $98.1\%$, a stability rate of $99.3\%$ and an average angular error of $5.2\degree$ on previously unseen objects. Compared to the $98.4\%$ success rate of \textbf{ITR} on seen objects (see Table~\ref{table:architecture}), these results suggest that the full \textbf{ITR-Q} algorithm can successfully generalize to unseen objects with only a minimal drop in performance.

\begin{figure}[t!]
\centering
\includegraphics[width=0.6\linewidth]{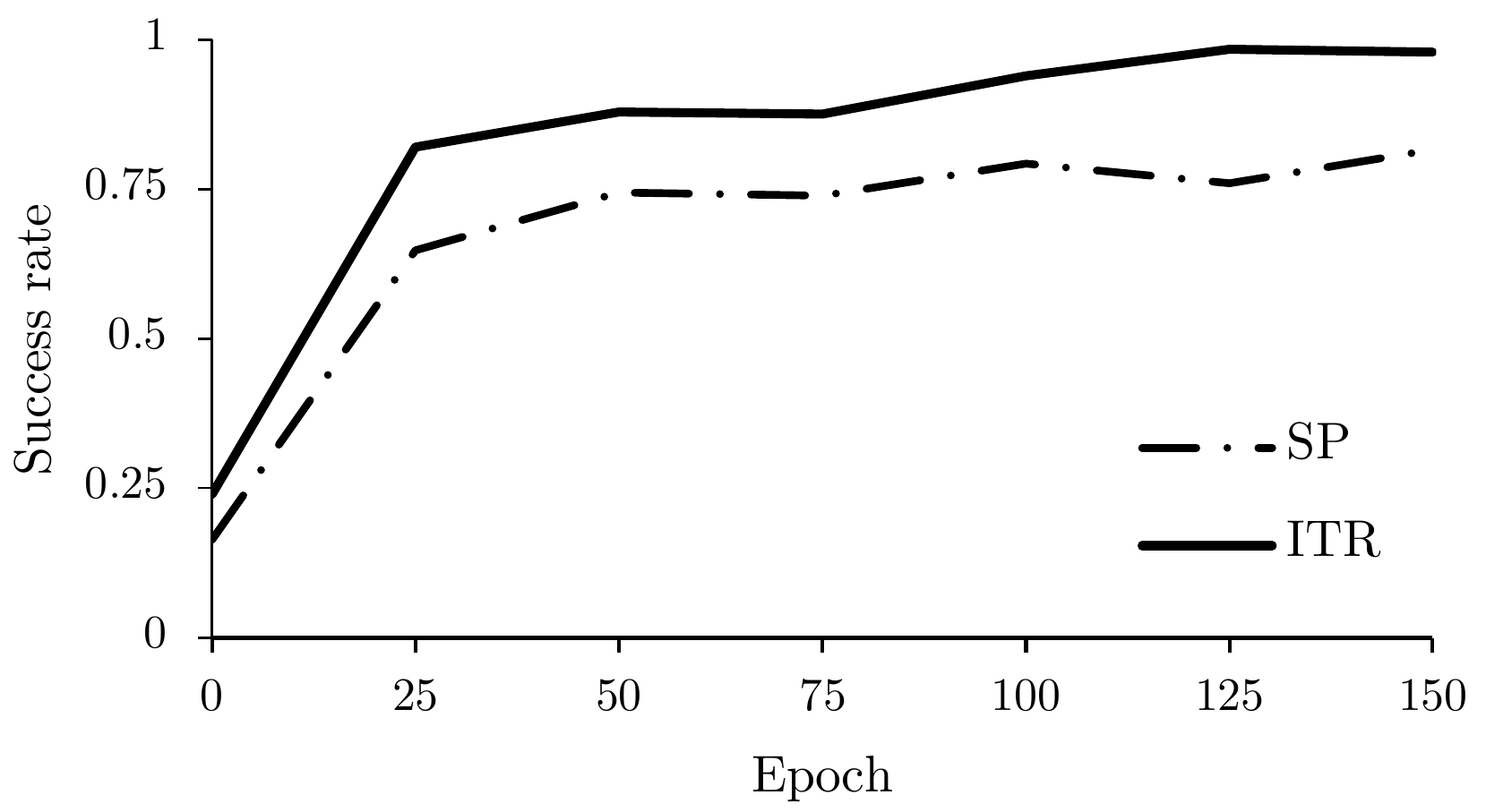}
\caption{\textbf{ITR} consistently outperformed \textbf{SP} during the training of PR-CNN.}
\label{traineval}
\end{figure}

The benefit of an iterative approach to using PR-CNN is validated by the $11.7$ percentage point improvement in success rate between \textbf{SP} and \textbf{ITR} methods. Moreover, the consistent outperformance of \textbf{ITR} compared to \textbf{SP} was observed throughout training, as seen in Figure~\ref{traineval}. This behavior suggests that the iterative approach can converge to a solution even if the underlying network has not been fully trained. 

\subsection{Common Failure Modes}
\begin{figure}[t!]
\centering
\includegraphics[width=0.8\linewidth]{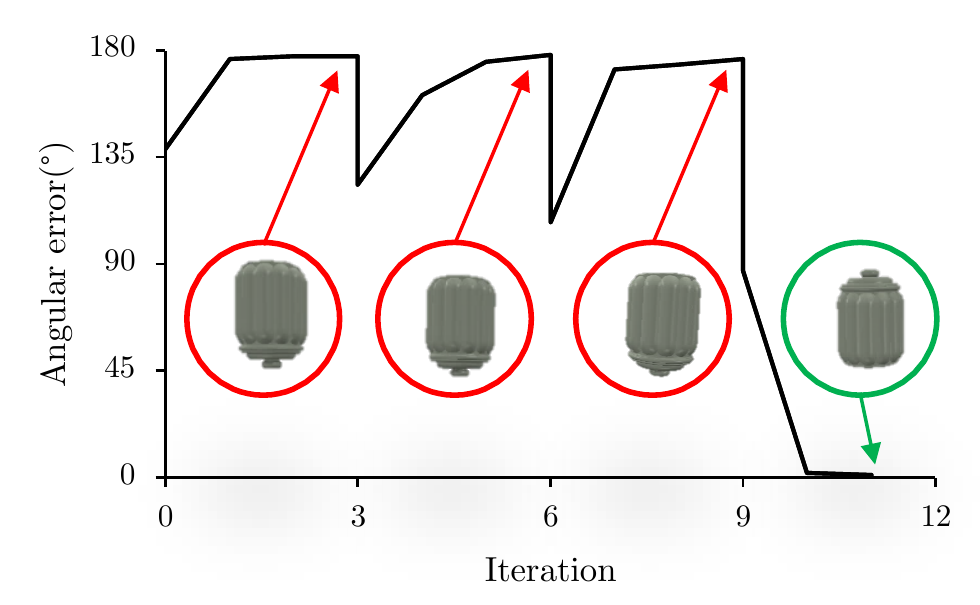}
\caption{Angular error of the cookie jar object is plotted at each iteration for \textbf{ITR-Q}. Red arrows point to iterations corresponding to when \textbf{ITR-Q} has identified low-quality convergence, and samples a new random orientation to seek a different final orientation. The final iteration highlighted in green is identified to be of a high quality by PQ-CNN, thus concluding the iterations.}
\label{fig:qualitydescent}
\end{figure}

The main failure mode was placing objects upside-down, particularly objects with relatively symmetrical features at either orientation, such as the cookie jar object (as seen in Figure~\ref{fig:qualitydescent}). This was particularly an issue for \textbf{SP} and \textbf{ITR} approaches if PR-CNN converged to this incorrect orientation. \textbf{ITR-Q} resolves this issue by identifying when a problematic convergence has occurred by setting a new randomized initial orientation, analogous to common optimization techniques used to escape from local minima. An example of this is illustrated in Figure~\ref{fig:qualitydescent}. For the cookie jar object, \textbf{ITR} had a success rate of $56.0\%$, while \textbf{ITR-Q} improved the success rate of the cookie jar object to $96.1\%$. This highlights the effectiveness of our full approach. The performance improvement on the cookie jar and similarly ambiguous objects make up most of the 2.0 percentage point improvement in overall success rate between \textbf{ITR} and \textbf{ITR-Q}.

\subsection{Generalization between object categories}
We observed that PR-CNN generalized well to objects of the same semantic categories present in the training set. Furthermore, during preliminary testing, when PR-CNN was trained on only bottles, the network would generalize well on other bottle like objects. However, PR-CNN trained on the single semantic category, bottles, did not generalize well to completely new object categories such as plates.

To mitigate this limitation, we trained our network on a wide variety of objects from different semantic categories, which allowed PR-CNN to generalize a lot better to new objects. Within the $50$ objects we have selected, we observed no performance degradation  increasing the number of object categories, apart from requiring more training data and more training time. However, further exploration of a more diverse object category pool and generalizing between them is an interesting direction for future work.

\section{Placement with a Simulated Robot}
\label{sec:simulated_robot}


We performed additional experiments with our trained networks by introducing a robotic arm. Rather than retraining the networks, we use Pix2Pix~\cite{isola2018imagetoimage} to segment the object in the image, removing the gripper. We then use \textbf{ITR-Q} on the segmented image. Assuming the gripper and grasped object act as a single rigid body (i.e. no slipping), we can execute object rotations by applying the same rotation to the robotic gripper. Placement is achieved by lowering the end effector until contact is detected via force sensing of the robot joints. The gripper fingers are then opened, and the end-effector is retracted along the reverse direction of the end-effector orientation. To increase the chances of solving the inverse kinematics, objects are placed on a surface elevated from the ground.


\subsection{Object Segmentation}

\begin{figure}[t]
\centering
\includegraphics[width=.25\linewidth]{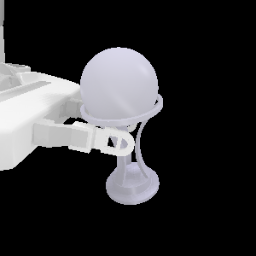}
\includegraphics[width=.25\linewidth]{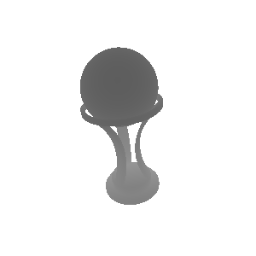}
\includegraphics[width=.25\linewidth]{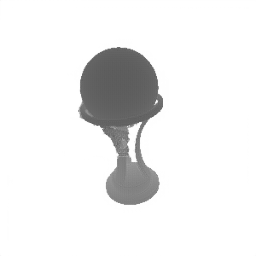}
\caption{Left: RGB component of the RGB-D input to the network. Middle: ground truth depth image. Right: Output of Pix2Pix which segmented out the gripper and hallucinated the gaps behind the fingers.}
\label{fig:pix2pix}
\end{figure}

Learning-based approaches (such as~\cite{florence2020robotsupervised}) have previously been proposed to segment out grippers from real scenes. However, these approaches do not reconstruct occluded pixels behind the gripper. We use Pix2Pix~\cite{isola2018imagetoimage} to segment out the object and hallucinate pixels occluded by robotic fingers, similar to~\cite{chen2020semantic}. The input to Pix2Pix is an RGB-D image with the gripper grasping the object. The ground truth is a depth image with only the object (see Fig.~\ref{fig:pix2pix}). We collected a dataset of $1,500$ images, $30$ for each object. We train a single network to remove the gripper from all viewpoints. We train Pix2Pix for $40$ epochs on the training set objects. We use the default parameters and loss function of the original model~\cite{isola2018imagetoimage}. The average network inference time was $21.09$ms for Pix2Pix (one RGB-D input).

\subsection{End Effector Constraints}

\begin{figure}[t]
\centering
\includegraphics[width=0.7\linewidth]{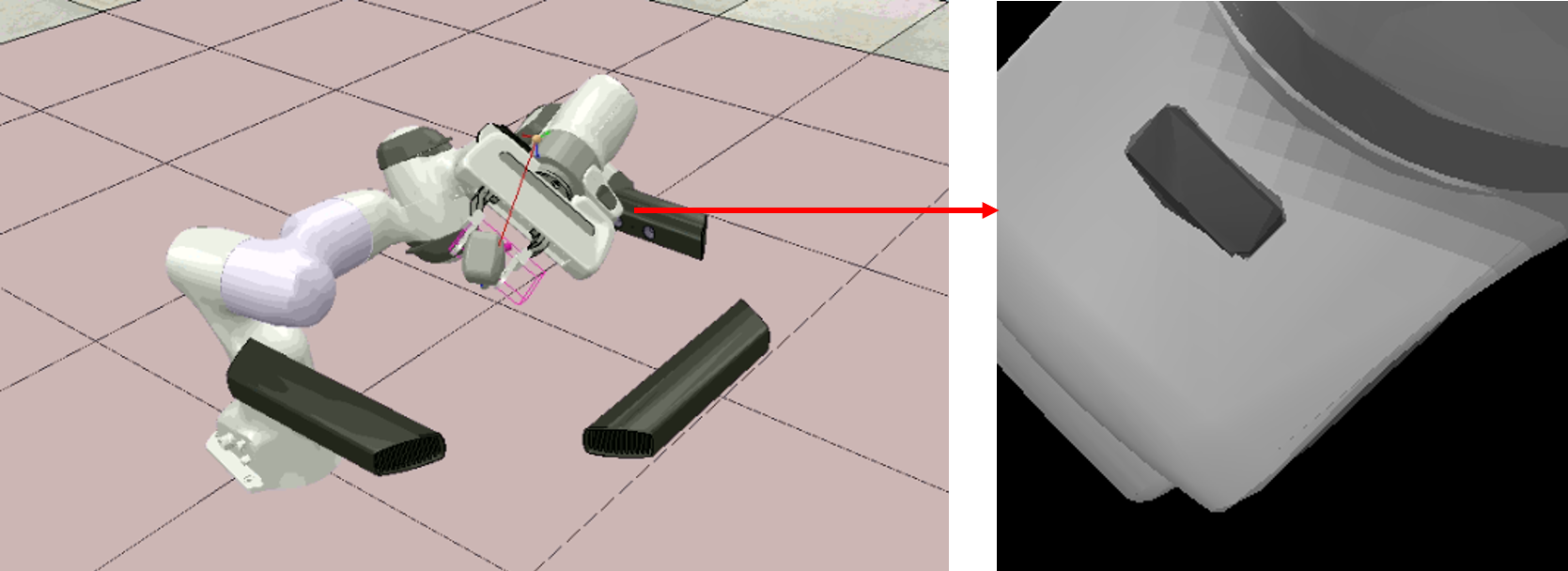}\\[0.5em]
\includegraphics[width=0.7\linewidth]{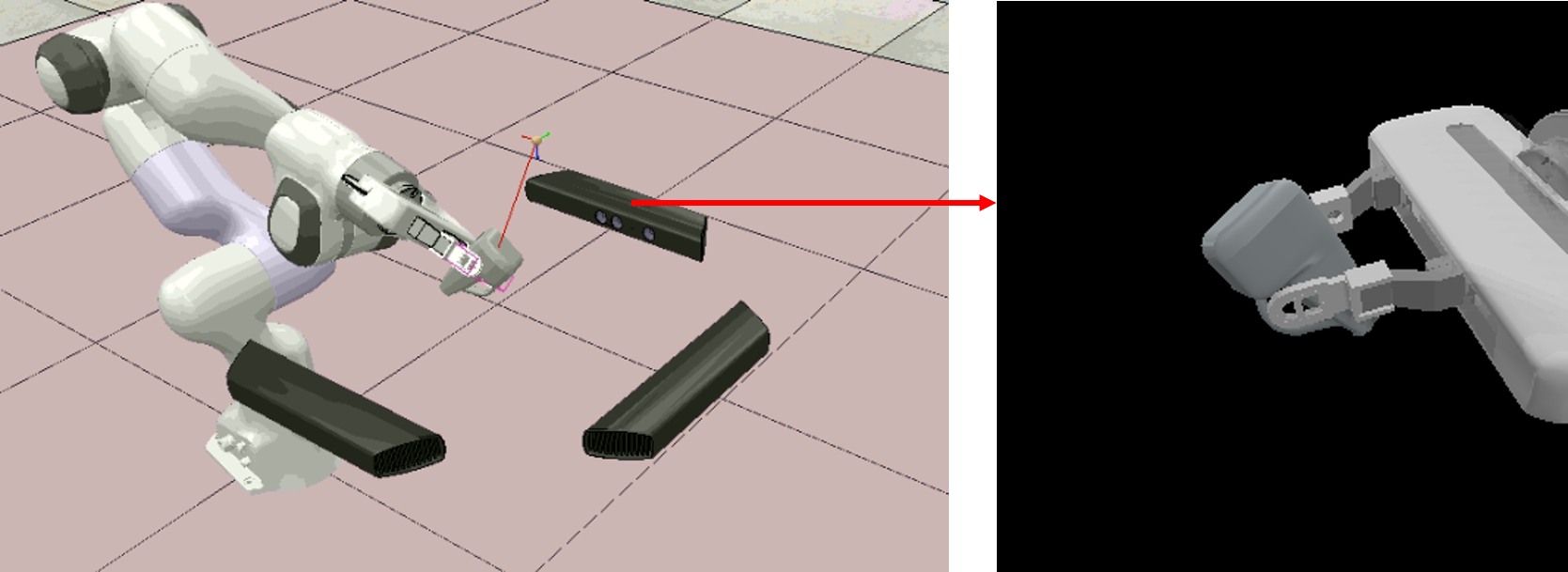}
\caption{Directly executing the rotations would cause the end-effector to occlude the cameras (Top), therefore, we apply a transformation to the output rotation around the stable axis to minimize occlusions (Bottom)}
\label{fig:Occluded}
\end{figure}

As we take an iterative approach, we cannot present the object to the cameras at a fixed gripper angle. Executing the rotations directly from PR-CNN would likely cause heavy occlusions in the images due to the gripper. See Fig.~\ref{fig:Occluded} (Top) for an example where the gripper almost entirely blocks one of the cameras. Therefore, we take a rotation of the gripper around the stable axis (defined as the z-axis in our problem) such that we minimize occlusions due to the gripper while still allowing for enough degrees of freedom to rotate the upright vector of the object to any orientation. After taking this action, the images become less occluded, as shown in Fig.~\ref{fig:Occluded} (Bottom).

\subsection{Results}


The results are evaluated on the same five test sets as Section~\ref{sec:unseen} and are shown in Table~\ref{table:gripper_results}. Compared to the previous experiment without the robot arm, the performance was lower as expected. The success rate decreased by $7.8$ percentage points and the stability rate decreased by $4.0$ percentage points. This is due to the inherent difficulty of the scenario, which includes the robot hand being in the images and the contact physics between the end-effector and the object. However, our approach can place previously unseen objects in stable orientations $95.3\%$ of the time and a success rate over $90\%$.

\begin{table}[h!]
\centering
\begin{tabular*}{\columnwidth}{@{\extracolsep{\stretch{1}}}*{4}{c}@{}}
\toprule
 & \textbf{\begin{tabular}[c]{@{}c@{}}Success\\ Rate (\%)\end{tabular}} & \textbf{\begin{tabular}[c]{@{}c@{}}Stability\\ Rate (\%)\end{tabular}} & \textbf{\begin{tabular}[c]{@{}c@{}}Angular\\ Error (\degree)\end{tabular}} \\ \midrule
\multicolumn{1}{l|}{\textbf{\textbf{ITR-Q} w/ Gripper}} & 90.3 $\pm$ 2.6 & 95.3 $\pm$ 1.9 & 16.7 $\pm$ 7.2 \\ \bottomrule
\end{tabular*}
\caption{Performance of \textbf{ITR-Q} in the presence of a simulated gripper. The same training and testing objects from Section~\ref{sec:unseen} were used. Despite the gripper introducing complexities into the input image, the algorithm is still able to achieve relatively high success rates. Standard Deviations are between the metrics obtained from the five test sets.}
\label{table:gripper_results}
\end{table}

The main failure mode of this situation is Pix2Pix\cite{isola2018imagetoimage} being unable to handle severe occlusions for small objects. In these cases, less useful information is provided to the networks, which cannot produce useful outputs. This problem could be alleviated by re-grasping the object.

\section{Placement with a Real Robot}
\label{sec:real_robot}


To demonstrate the feasibility of our proposed approach and the potential for sim-to-real transfer, we implemented our object placement approach on a Franka Emika Panda robotic arm using a single Kinect V2 depth camera. To allow for performance comparisons with the previous experiment involving a single camera, the \textbf{ITR} approach was used for the real-world experiments. We evaluated our approach in a pick and place scenario, where an object starts on a table and has to be placed onto a shelf. The system consists of two computers connected via TCP/IP.  Multiple computers were used due to the high USB bandwidth required for running multiple RGB-D cameras. Each computer uses the Melodic version of the Robot Operating System running on Ubuntu 18.04 LTS operating system. 





\begin{figure}[t]
\centering
    \includegraphics[trim=2cm 0.5cm 1cm 7.5cm, 
clip, width=0.6\linewidth]{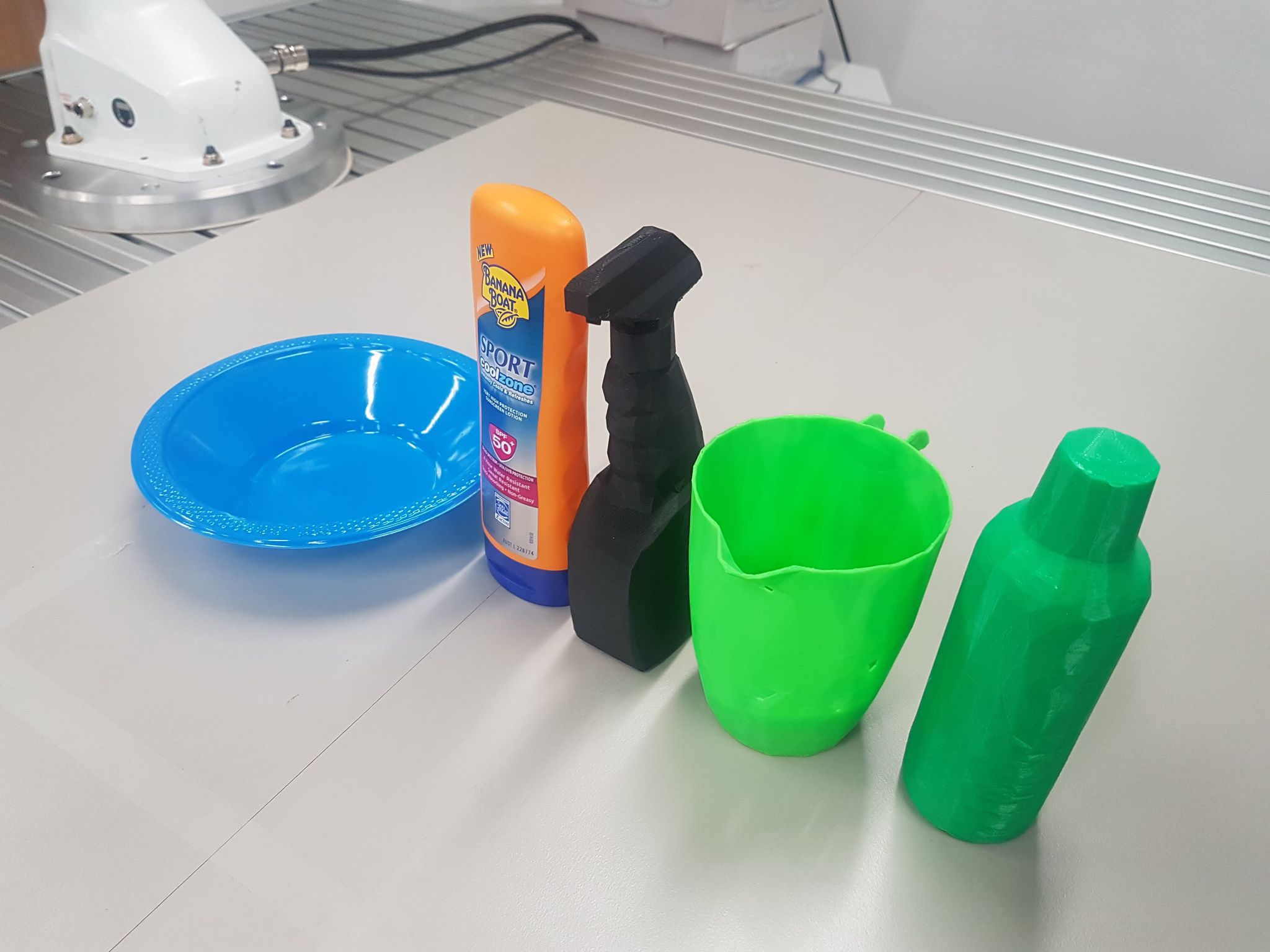}
\caption{The 5 objects used for real robot experiments. Left to right: Bowl, Sunscreen, 3D printed Spray Bottle, 3D printed Green Pitcher and 3D printed Green Bottle. Only the Green Bottle was in the original 50 training objects.}
\label{fig:real-objects}
\end{figure}


To generate realistic object grasps for the algorithm to consider placement, we use GGCNN~\cite{morrison2018closing} to generate viable grasp points from a given depth image generated by a wrist-mounted RealSense D435 RGB-D camera. Once the object is grasped, the arm moves to a predefined location in front of the Kinect V2.


Some modifications were made to the original approach to increase the likelihood of success in real-world experiments. (1) The Kinect V2 images were pre-processed before being used in the \textbf{ITR} algorithm, which involved removing the background of the depth and color images based on a minimum and maximum depth threshold. (2) We consider the output of PR-CNN for a moving window of depth images (window size = 5). If the output of all $5$ frames is within $10\degree$, the average rotation across the window is executed. This helps remove possible outliers from the output. If this does not occur within $100$ frames, the window with the lowest variance is used. (3) We controlled the arm with the Manipulability Motion Controller (MMC)~\cite{haviland2020maximising}, a controller designed to maximize the manipulability of the arm. This was chosen to avoid joint limits when executing complex rotations with the arm. (4) When placing the object at the predetermined position on the shelf, the target end-effector pose would often correspond to kinematically infeasible joint configurations. Hence, we randomly sampled positions about the shelf and angles around the stable axis to find a feasible placement pose. This maximized the possibility of successfully finding placement positions within the joint limits of the robotic arm.



We ran ten trials of five different objects (shown in Figure~\ref{fig:real-objects}). One object was 3D printed from the data set. Four objects were not featured in the training set of the networks. Objects were chosen which could be segmented using Pix2Pix, as we are focusing on testing the approach to placing objects.

\subsection{Results}

Overall, we achieved an $88.0\%$ success rate in real-world experiments. Despite the irregularities introduced in the real world, the results show that we can achieve results comparable to those we achieved in simulation. The network performed very well with the Bowl, however, this success rate is skewed, as the bowl was likely to fall into the correct orientation once placed.

\begin{table}[h!]
\centering
\begin{tabular*}{\columnwidth}{@{\extracolsep{\stretch{1}}}*{3}{c}@{}}
\toprule
 & \multicolumn{1}{l}{\textbf{Success Rate (\%)}} & \multicolumn{1}{l}{\textbf{Avg. Num. Iterations}} \\ \midrule
\multicolumn{1}{l|}{\textbf{Bowl}} & 100.0 & 1.8 \\
\multicolumn{1}{l|}{\textbf{Sunscreen}} & 90.0 & 1.8 \\
\multicolumn{1}{l|}{\textbf{Spray Bottle}} & 80.0 & 1.6 \\
\multicolumn{1}{l|}{\textbf{Green Pitcher}} & 80.0 & 1.7 \\
\multicolumn{1}{l|}{\textbf{Green Bottle}} & 90.0 & 1.3 \\ \bottomrule
\end{tabular*}
\caption{Real-world performance of \textbf{ITR using a single depth camera, grasping objects from the table and placing the object on to a shelf. The results are comparable to those achieved in the simulated experiment ($85.2\%$).}}
\label{tab:my-table}
\end{table}

\subsection{Failure Modes}


The most common failure was due to kinematic constraints, where the robotic arm could not successfully execute the orientation requested by the algorithm. This was partially mitigated by the introduction of MMC~\cite{haviland2020maximising}. This could be improved further by considering motion planning for this application, such as the work done by \cite{holladay}.

Moreover, the robot would sometimes occlude the object from the singular camera . However, this is a known issue, and as discussed in Section~\ref{sec:design-parameteres}, we expect the use of additional cameras to improve the performance of PR-CNN.

Limitations due to direct sim-to-real transfer also introduced errors, which is an open problem in robot learning~\cite{sim2real}. Inaccurate segmentation of the object from the gripper using the Pix2Pix network and noise from the images captured by the Kinect V2 were observed during the experiments. Future work to investigate and mitigate the effects of these artifacts is expected to improve the performance of the algorithms further.





\section{Conclusion and Future Work}
\label{sec:conclusion}

In this work, we propose an approach to rotate grasped objects into orientations to be placed in stable, upright orientations. We show the feasibility of learning to place objects from depth images without object classification or explicit pose estimation. Our simulation experiments suggest that our iterative approach \textbf{ITR-Q} performs better than placing objects on their largest supporting planes. The proposed approach is general enough so that our two proposed neural networks can be replaced by alternative approaches, however, we show that learning-based approaches provide the opportunity to learn human-annotated upright orientations, giving an edge over analytical methods. Our work also shows potential for sim-to-real transfer learning and justifies the need for more research to generalize the approach to different object classes and shapes.


Our current iterative approach only re-evaluates the object's orientation after it has completed the rotation, making it slow to react to disturbances. We reserve a closed-loop reactive approach as future work, analogous to grasping in~\cite{morrison2018closing}, that can re-evaluate the rotation at every time step, and hence will be more effective in dealing with object slip and disturbances. A limitation of our approach is that it is designed for objects with a single defined preferred orientation, however, some objects such as cans and boxes have multiple preferred orientations. It is possible to extend our representation to include such objects by considering multiple ground truth rotations. Another interesting idea for future work is further exploring the ability to estimate the quality of placement. This would allow exploring more creative placement methods such as controlled dropping of an object~\cite{holladay} in a feasible configuration of the arm.


\bibliographystyle{IEEEtran}
\bibliography{refs}

\begin{thebibliography}{10}
\providecommand{\url}[1]{#1}
\csname url@samestyle\endcsname
\providecommand{\newblock}{\relax}
\providecommand{\bibinfo}[2]{#2}
\providecommand{\BIBentrySTDinterwordspacing}{\spaceskip=0pt\relax}
\providecommand{\BIBentryALTinterwordstretchfactor}{4}
\providecommand{\BIBentryALTinterwordspacing}{\spaceskip=\fontdimen2\font plus
\BIBentryALTinterwordstretchfactor\fontdimen3\font minus
  \fontdimen4\font\relax}
\providecommand{\BIBforeignlanguage}[2]{{%
\expandafter\ifx\csname l@#1\endcsname\relax
\typeout{** WARNING: IEEEtran.bst: No hyphenation pattern has been}%
\typeout{** loaded for the language `#1'. Using the pattern for}%
\typeout{** the default language instead.}%
\else
\language=\csname l@#1\endcsname
\fi
#2}}
\providecommand{\BIBdecl}{\relax}
\BIBdecl

\bibitem{DataDrivenGrasp}
J.~{Bohg}, A.~{Morales}, T.~{Asfour}, and D.~{Kragic}, ``Data-driven grasp
  synthesis—a survey,'' \emph{IEEE Trans. Robot.}, 2014.

\bibitem{lenz2013deep}
I.~Lenz, H.~Lee, and A.~Saxena, ``Deep learning for detecting robotic grasps,''
  \emph{IJRR}, 2015.

\bibitem{mahler2017dexnet}
J.~Mahler, J.~Liang, S.~Niyaz, M.~Laskey, R.~Doan, X.~Liu, J.~A. Ojea, and
  K.~Goldberg, ``Dex-net 2.0: Deep learning to plan robust grasps with
  synthetic point clouds and analytic grasp metrics,'' in \emph{RSS}, 2017.

\bibitem{pinto2015supersizing}
L.~Pinto and A.~Gupta, ``Supersizing self-supervision: Learning to grasp from
  50k tries and 700 robot hours,'' in \emph{IEEE ICRA}, 2016.

\bibitem{morrison2018closing}
D.~Morrison, P.~Corke, and J.~Leitner, ``{Closing the Loop for Robotic
  Grasping: A Real-time, Generative Grasp Synthesis Approach},'' in
  \emph{Robotics: Science and Systems (RSS)}, 2018.

\bibitem{2018Garrett}
C.~R. Garrett, T.~Lozano-Pérez, and L.~P. Kaelbling, ``Ffrob: Leveraging
  symbolic planning for efficient task and motion planning,'' \emph{IJRR},
  2018.

\bibitem{Jiang2011}
Y.~Jiang, C.~Zheng, M.~Lim, and A.~Saxena, ``Learning to place new objects,''
  \emph{IEEE ICRA}, 2011.

\bibitem{harada2012}
K.~Harada, T.~Tsuji, K.~Nagata, N.~Yamanobe, H.~Onda, T.~Yoshimi, and Y.~Kawai,
  ``Object placement planner for robotic pick and place tasks,'' in
  \emph{IEEE/RSJ IROS}, 2012.

\bibitem{haustein2019object}
J.~A. {Haustein}, K.~{Hang}, J.~{Stork}, and D.~{Kragic}, ``Object placement
  planning and optimization for robot manipulators,'' in \emph{IROS}, 2019.

\bibitem{Fu2008}
H.~Fu, D.~Cohen-Or, G.~Dror, and A.~Sheffer, ``Upright orientation of man-made
  objects,'' in \emph{ACM SIGGRAPH}, 2008.

\bibitem{EdSigner}
A.~{Edsinger} and C.~C. {Kemp}, ``Manipulation in human environments,'' in
  \emph{IEEE-RAS International Conference on Humanoid Robots}, 2006.

\bibitem{baumgartl2013}
J.~{Baumgartl}, P.~{Kaminsky}, and D.~{Henrich}, ``A geometrical placement
  planner for unknown sensor-modelled objects and placement areas,'' in
  \emph{IEEE ROBIO}, 2013.

\bibitem{baumgartl2014c}
J.~Baumgartl, T.~Werner, P.~Kaminsky, and D.~Henrich, ``A fast, gpu-based
  geometrical placement planner for unknown sensor-modelled objects and
  placement areas,'' in \emph{IEEE ICRA}, 2014.

\bibitem{HARADA20141463}
K.~Harada, T.~Tsuji, K.~Nagata, N.~Yamanobe, and H.~Onda, ``Validating an
  object placement planner for robotic pick-and-place tasks,'' \emph{Rob Auton
  Syst}, 2014.

\bibitem{Jiang2012}
Y.~Jiang, M.~Lim, C.~Zheng, and A.~Saxena, ``Learning to place new objects in a
  scene,'' \emph{IJRR}, 2012.

\bibitem{Paolini2014}
R.~Paolini, A.~Rodriguez, S.~S. Srinivasa, and M.~T. Mason, ``A data-driven
  statistical framework for post-grasp manipulation,'' \emph{IJRR}, 2014.

\bibitem{manuelli2019kpam}
L.~Manuelli, W.~Gao, P.~Florence, and R.~Tedrake, ``kpam: Keypoint affordances
  for category-level robotic manipulation,'' in \emph{ISRR}, 2019.

\bibitem{gao2019kpamsc}
W.~Gao and R.~Tedrake, ``kpam-sc: Generalizable manipulation planning using
  keypoint affordance and shape completion,'' 2019.

\bibitem{mitash2020task}
C.~Mitash, R.~Shome, B.~Wen, A.~Boularias, and K.~Bekris, ``Task-driven
  perception and manipulation for constrained placement of unknown objects,''
  \emph{IEEE RA-L}, vol.~5, no.~4, pp. 5605--5612, 2020.

\bibitem{carreira2016human}
J.~Carreira, P.~Agrawal, K.~Fragkiadaki, and J.~Malik, ``Human pose estimation
  with iterative error feedback,'' in \emph{IEEE CVPR}, 2016.

\bibitem{actorcritic}
T.~Haarnoja, A.~Zhou, K.~Hartikainen, G.~Tucker, S.~Ha, J.~Tan, V.~Kumar,
  H.~Zhu, A.~Gupta, P.~Abbeel, and S.~Levine, ``Soft actor-critic algorithms
  and applications,'' 2019.

\bibitem{he2015deep}
K.~He, X.~Zhang, S.~Ren, and J.~Sun, ``Deep ` learning for image recognition,''
  in \emph{IEEE CVPR}, 2016.

\bibitem{Deng09imagenet}
J.~Deng, W.~Dong, R.~Socher, L.~jia Li, K.~Li, and L.~Fei-fei, ``Imagenet: A
  large-scale hierarchical image database,'' in \emph{CVPR}, 2009.

\bibitem{openai2019learning}
OpenAI, ``Learning dexterous in-hand manipulation,'' \emph{The International
  Journal of Robotics Research}, vol.~39, no.~1, pp. 3--20, 2020.

\bibitem{zhou2018continuity}
Y.~Zhou, C.~Barnes, J.~Lu, J.~Yang, and H.~Li, ``On the continuity of rotation
  representations in neural networks.'' in \emph{CVPR}, 2019.

\bibitem{Chen2013TheCA}
Z.~Chen and F.~Cao, ``The construction and approximation of neural networks
  operators with gaussian activation function,'' in \emph{Math. Comm.}, 2013.

\bibitem{coumans2010bullet}
E.~Coumans, ``Bullet physics engine,'' \emph{http://bulletphysics. org}, 2010.

\bibitem{pyrep}
S.~James, M.~Freese, and A.~J. Davison, ``Pyrep: Bringing v-rep to deep robot
  learning,'' \emph{arXiv preprint arXiv:1906.11176}, 2019.

\bibitem{Zhou2018}
Q.-Y. Zhou, J.~Park, and V.~Koltun, ``{Open3D}: {A} modern library for {3D}
  data processing,'' \emph{arXiv:1801.09847}, 2018.

\bibitem{isola2018imagetoimage}
P.~Isola, J.~Zhu, T.~Zhou, and A.~A. Efros, ``Image-to-image translation with
  conditional adversarial networks,'' in \emph{IEEE CVPR}, 2017.

\bibitem{florence2020robotsupervised}
V.~{Florence}, J.~J. {Corso}, and B.~{Griffin}, ``Robot-supervised learning for
  object segmentation,'' in \emph{2020 ICRA}, 2020, pp. 1343--1349.

\bibitem{chen2020semantic}
Z.~Chen, D.~Ting, R.~Newbury, and C.~Chen, ``Semantic segmentation for
  partially occluded apple trees based on deep learning,'' \emph{Comput.
  Electron. Agric.}, 2021.

\bibitem{haviland2020maximising}
J.~Haviland and P.~Corke, ``A purely-reactive manipulability-maximising motion
  controller,'' \emph{arXiv:2002.11901}, 2020.

\bibitem{holladay}
A.~{Holladay}, J.~{Barry}, L.~P. {Kaelbling}, and T.~{Lozano-Pérez}, ``Object
  placement as inverse motion planning,'' in \emph{IEEE ICRA}, 2013.

\bibitem{sim2real}
F.~{Sadeghi}, A.~{Toshev}, E.~{Jang}, and S.~{Levine}, ``Sim2real viewpoint
  invariant visual servoing by recurrent control,'' in \emph{CVPR}, 2018.

\end{thebibliography}

\end{document}